\documentclass[letterpaper]{article} 
\usepackage[submission]{aaai23}  
\usepackage{times}  
\usepackage{helvet}  
\usepackage{courier}  
\usepackage[hyphens]{url}  
\usepackage{graphicx} 
\usepackage{natbib}  
\usepackage{caption} 
\usepackage{algorithm}
\usepackage{algorithmic}
\usepackage{color}
\usepackage{amsmath}
\usepackage{booktabs}

\usepackage{newfloat}
\usepackage{listings}
\DeclareCaptionStyle{ruled}{labelfont=normalfont,labelsep=colon,strut=off} 
\floatstyle{ruled}
\newfloat{listing}{tb}{lst}{}
\floatname{listing}{Listing}
\title{Prototype Fission: Closing Set for Robust Open-set Semi-supervised Learning}

\author{
    Xuwei Tan\textsuperscript{\rm 1,2,\footnote{Work completed during an internship at OPPO Research Institute.}},
    Yi-Jie Huang \textsuperscript{\rm 1},
    Yaqian Li \textsuperscript{\rm 1,\footnote{Corresponding author.}}
}

\affiliations{
    \textsuperscript{\rm 1} OPPO Research Institute\\
    \textsuperscript{\rm 2} Department of Computer Science and Engineering, The Ohio State University\\[0.5ex]
    \texttt{tan.1206@osu.edu}, 
    \texttt{huangyijie@oppo.com}, 
    \texttt{liyaqian@oppo.com}
}

\usepackage{bibentry}

\begin{document}

\maketitle

\begin{abstract}

Semi-supervised Learning (SSL) has been proven vulnerable to out-of-distribution (OOD) samples in realistic large-scale unsupervised datasets due to over-confident pseudo-labeling OODs as in-distribution (ID). A key underlying problem is class-wise latent space spreading from closed seen space to open unseen space, and the bias is further magnified in SSL's self-training loops. To close the ID distribution set so that OODs are better rejected for safe SSL, we propose Prototype Fission~(PF) to divide class-wise latent spaces into compact sub-spaces by automatic fine-grained latent space mining, driven by coarse-grained labels only. Specifically, we form multiple unique learnable sub-class prototypes for each class, optimized towards both diversity and consistency. The Diversity Modeling term encourages samples to be clustered by one of multiple sub-class prototypes, while the Consistency Modeling term clusters all samples of the same class to a global prototype. Instead of "opening set", i.e., modeling OOD distribution, Prototype Fission "closes set" and makes it hard for OOD samples to fit in sub-class latent space. Therefore, PF is compatible with existing methods for further performance gains. Extensive experiments validate the effectiveness of our method in open-set SSL settings in terms of successfully forming sub-classes, discriminating OODs from IDs and improving overall accuracy. Codes will be released.

\end{abstract}

\section{Introduction}\label{sec:introduction}
\begin{figure*}[ht]
	\centering
	\centerline{\includegraphics[width=16cm]{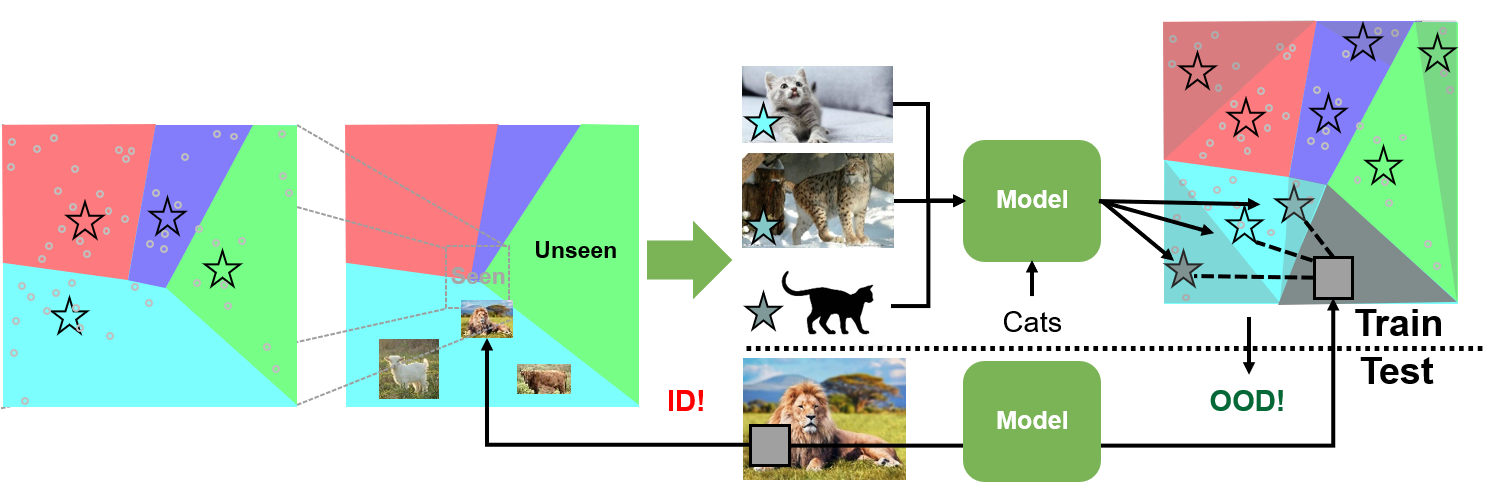}}
	\vspace{-0.0cm}
  	\caption{Motivation: Due to the local-optimal that hyper-planes trained on seen space already correctly classify seen samples, there is no need for learning further compactness. However, catastrophic ID-OOD confusions happen in the unseen space. Our Prototype Fission instead further casts consistency and diversity to properly divide these spaces, thereby OOD samples can be better rejected. }
	\label{fig:Motivation}
\end{figure*}

{D}{eep} Learning has achieved remarkable success in the past years, but the hunger for data has always been the inherent bottleneck of deep models. Promising free lunch derived from large-scale unlabeled samples, Semi-supervised Learning~(SSL) has achieved record-breaking data-efficiency in different computer vision tasks~\cite{sohn2020fixmatch, xu2021end} and attracted broad attention.

By self-training on pseudo-labeled unlabeled data, the nature of SSL is to magnify confirmation, regardless of correctness or bias. Different from frequently used closed-set datasets, real-world open-set datasets present open class semantics, mismatched long-tail effects, and unknown domain gaps, among which the out-of-distribution (OOD) class semantics is a harmful and hard-to-solve factor in terms of magnifying ID vs OOD confusion. In the SSL self-training loop, many OOD samples are over-confidently pseudo-labeled as ID samples and the bias is further magnified in the following iterations. To tackle this problem, many researchers focus on compacting intra-class latent space while distancing inter-class latent space~\cite{yang2018robust, lu2022pmal, shu2020p}, adaptively re-weighting unlabeled samples~\cite{guo2020safe}, finding reciprocal points~\cite{chen2020learning, chen2021adversarial} and clustering unknown classes~\cite{cao22} for better OOD latent space separation. Despite their certain success, celebrating the performance gains while enabling safety on open-set unsupervised datasets remains the Holy Grail for this line of SSL research~\cite {guo2020safe}. 

Besides the widely discussed valuable awareness of unseen latent space~\cite{chen2020learning, chen2021adversarial, cao22}, the confirmation bias, i.e. over-confident pseudo-labeling, is also deeply rooted in the internal distribution deviation of ID classes. As is illustrated in Fig.\ref{fig:Visualization}, the existence of multiple sub-semantics can be easily observed from the sample distribution that shares an identical class label, e.g.., the definition of 'horse' can be too coarse-grained and can be further divided into 'head of a horse', 'human riding a horse' and 'the full-length portrait of a horse', which have significantly different appearances while the high-level agreement is shared. In worse cases, the class-wise semantic can be formed by sub-optimally mixing different concepts by human annotators or large-scale vision-language pairing~\cite{radford2021learning}. In such cases, clustering all samples to one prototype makes the class-wise latent space rather diffused despite the effort on compacting the space: as a consequence, OOD samples can easily fit in and further diffuse the space, as is illustrated in Fig.~\ref{fig:Motivation}.

To lift this underlying conflict, in this paper, we propose Prototype Fission which employs multiple learnable prototypes to divide diffused class-wise latent spaces into multiple compact sub-class spaces. Specifically, (1) A Global Prototype is used to cluster all data of a class to a single center. (2) Multiple Local Prototypes are used to cluster different subsets of a given class. (3) Global Prototype and Local Prototypes are combined in an adaptively weighted way. (4) An Adversarial Prototype Fission training strategy is proposed to maintain both the diversity and consistency of the aforementioned prototypes. The Diversity Modeling term encourages samples to be assigned to different sub-class prototypes, while the Consistency Modeling term clusters all samples of the same class to a global prototype. An initial result is also illustrated in Fig.\ref{fig:Visualization}: the sub-semantics are mined by our Prototype Fission model given only the coarse-grained class label. A further merit of Prototype Fission is its clear focus on ID space: it can be combined with multiple safe SSL methods focusing on OOD sample modeling. In summary, the major achievements of this paper are as follows:

\begin{enumerate}
\item We proposed a framework with multiple learnable prototypes to make class-wise latent space more compact. An Adversarial Prototype Fission training strategy is proposed to maintain both the diversity and consistency of the aforementioned prototypes. Experiment results show that Prototype Fission effectively divides classes into sub-classes and better discriminates OODs from ID samples. 

\item Different from the 'opening set' methods modeling unseen space, i.e. clustering novel classes~\cite{cao22} or forming reciprocal points~\cite{chen2020learning, chen2021adversarial}, the proposed Prototype Fission focus on 'closing set', i.e., better modeling ID space. This nature makes it highly compatible with 'opening set' methods for further accuracy and safety gains. 



\end{enumerate}

\section{Related Works}\label{sec:related_works}

\subsection{Open-set \& Open-world SSL Frameworks} 

While open-world SSL additionally mines novel classes and performs the evaluation on novel classes afterward, an open-set \& open-world SSL framework is typically a combination of Feature Extractors, Prototype Learning~(PL), Metric Learning and Semi-supervised Learning tuned towards open-set \& open-world challenges. For example, the open-world SSL method ORCA~\cite{cao22}, employed SimCLR~\cite{chen2020simple} trained on open-distribution as the feature extractor, formed novel-class prototypes, and used cosine-based distance metric. OpenLDN~\cite{rizve2022openldn} further replaced the cosine-based metric with improved Metric Learning, $i.e.$, a learnable similarity predictor, and added an extra stage performing pseudo-closed-set SSL, which better enjoys the gain from data augmentation-based SSL methods. This trend of evolving from single-trick solutions to complicated frameworks motivates us to organize related works in a modularized way as follows:

\subsubsection{Open-set Feature Extraction:}
Formulating an OOD-proof feature extractor is the foundation of solving open-set tasks. ImageNet-pretraining~\cite{deng2009imagenet} can be seen as an initial attempt, yet it formed a limited number of semantic concepts based on a huge number of human annotations. Vision-based Self-supervised Learning~\cite{chen2020simple, he2022masked} form tasks to model inherit dependencies among images, $e.g.$ instance discrimination tasks or impainting-based tasks, to lift the demand on annotation. Given the rapid growth of web-sourced paired vision-language~(VL) data containing the unlimited number of semantics, VL-Pretrained models, $e.g.$ CLIP~\cite{radford2021learning}, are becoming popular open-set feature extractors.

In this paper, the feature extractor is set as a representative milestone, SimCLR~\cite{chen2020simple}, for fair comparison against existing open-set \& open-world SSL methods. We also show the effectiveness of replacing SimCLR with a more recent advance in this line of research, CLIP~\cite{radford2021learning}, on applicable datasets.

\begin{figure*}[ht]
	\centering
	\centerline{\includegraphics[width=16cm]{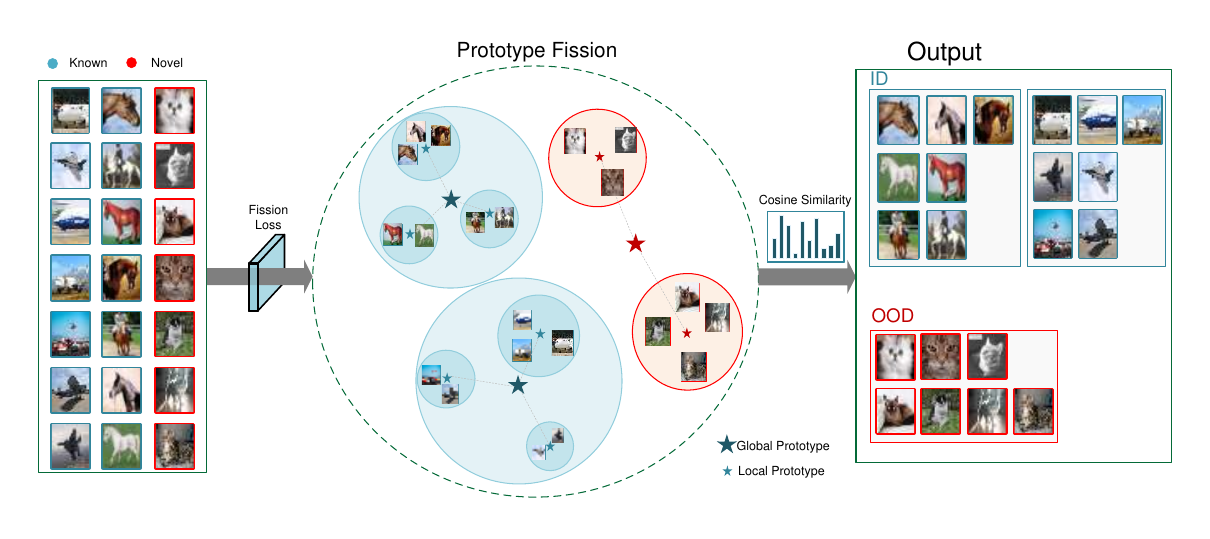}}
	\vspace{-0.0cm}
	\caption{Prototype Fission consists of a shared feature extractor $f$ and $V$ prototypes per class to mine sub-class semantics. The Diversity Modeling strategy and the Consistency Modeling strategy form adversarial objectives in that different prototypes are both concentrated in the same class-wise space and branched to different sub-spaces, thereby ID spaces are more fine-grained to reject OOD samples. }
	\label{fig:Framework}
\end{figure*}

\subsubsection{ID Prototype Learning:} PL methods~\cite{yang2018robust, shu2020p, lu2022pmal} normally perform pre-clustering and select images close to class centers as prototypes, or set prototypes as trainable vectors. Especially, PL methods are aware of the potential merit of using multiple prototypes. For example, CPL~\cite{yang2018robust} proposes forming multiple learnable prototypes per class. However, different numbers of prototypes made little difference in performance due to the lack of explicit regularization. By Monte-Carlo stability evaluation, PMAL~\cite{lu2022pmal} proposed to select multiple static prototype images featuring both diversity and density. Our proposed method fits in this line of research. Especially, the proposed method differs in forming multiple learnable prototypes per class with carefully designed constraints to be effective in Open-set Semi-supervised Learning tasks. 

\subsubsection{OOD Prototype Learning:} As reasonably formed OOD prototypes can serve as anchors pulling OOD samples from ID space, forming OOD prototypes makes a significant contribution to ID class robustness in addition to recognizing novel classes. As an initial step, Reciprocal Points Learning~\cite{chen2020learning, chen2021adversarial} proposed an abstract concept named Reciprocal Points to model per-class counter-prototypes. Further methods extend the given ID class list and form frameworks unifying OOD clustering and ID learning. As it is hard yet necessary to synchronize the two objectives during the training stage,  multiple advances tried in different ways: ORCA~\cite{cao22} slows down ID learning so that OOD clustering catches up, NACH~\cite{guorobust} adaptively accelerates OOD clustering, while OpenLDN~\cite{rizve2022openldn} form two-stage scheme to transfer open-world SSL into closed-set SSL to make the synchronization happen naturally and the extensive data augmentation more compatible.


\subsubsection{Metric Learning for OOD Detection:} In the open-set vision literature, a straightforward and widely discussed strategy is to detect OOD samples based on certain metrics and reject them in case ID-OOD confusion happens. The metric for OOD detection can be roughly divided into distance-based~\cite{park2021opencos,zhou2021step}, classifier-based~\cite{saito2021openmatch,huang2021trash}, and meta-learning-based~\cite{guo2020safe}. Different from the common usage of OOD detection and rejection for open-set robustness, the proposed Prototype Fission suppresses ID-OOD confusion by compacting ID latent spaces, which is orthogonal with Metric Learning methods, and can further benefit from them.

\section{Method}\label{sec:method}

To achieve safe and robust semi-supervised learning, we proposed Prototype Fission consisting of the formation of multiple sub-class prototypes and an adversarial training strategy for both intra-class consistency and intra-class diversity. 

\subsection{Multiple Prototype Formation}
As shown in Figure \ref{fig:Framework}, Prototype Fission consists of a shared feature extractor $f$ and $V$ prototypes per class to mine sub-class semantics, i.e., for $N$ classes explicitly defined by human annotators, we form $V\cdot N$ prototypes. Instead of using geometric centers of features as prototypes in many existing PL methods, we model our prototypes as normalized vectors implemented as head weights and calculate cosine similarity between prototypes and image features for the convenience of following operations. 

To maintain intra-class diversity and consistency, the $i$th prototype of class $c$, $p_{(c,i)}$ is explicitly designed to be a combination of a global consistency component $g_{c}$ shared among $V$ prototypes of class $c$, and a local diversity term  $l_{(c,i)}, i\in[1,..., V]$. Instead of manually tuning the contribution portion of global and local components, we form the combination of $g_{c}$ and $l_{(c,i)}$ as a sum weighted by a trainable parameter $\lambda_{c,i} \in \Lambda_{c} = [\lambda_{(c,1)},\lambda_{(c,2)},...,\lambda_{(c,V)}]$ as follows:
\begin{equation}
\begin{aligned}
    &p_{(c,i)} = (1 - \lambda_{(c,i)}) \cdot norm(g_{c}) +  \lambda_{(c,i)} \cdot norm(l_{(c,i)}),\\
    &where\  0<\lambda_{(c,i)}<1
\end{aligned}
\end{equation}
where we normalize $g_{c}$ and $l_{(c,i)}$ to make the weighting operation invariant to the magnitudes. 

To avoid box constraints $\lambda_{(c,i)} \in [0,1]$ on $\lambda_{(c,i)}$ and make the optimization  more effective, we follow \cite{carlini2017towards} and form an auxiliary variable $\delta$ to optimize $\lambda_{(c,i)}$ as follows:
\begin{equation}
\begin{split}
    {\lambda_{(c,i)}=\frac{1}{2}(\tanh{\delta_{(c,i)}}+1), \delta_{(c,i)} \in (-\infty,+\infty)}
\end{split}
\end{equation}

During training, the global component $g_{c}$ can be updated by every sample of class $c$, which models the general agreement among different prototypes. The local component $l_{(c,i)}$ further models the exclusive sub-class uniqueness. The weight term $\lambda_{(c,i)}$ adaptively balances the contribution of $g_{c}$ and $l_{(c,i)}$ by following the Adversarial Prototype Fission learning strategy. 


\subsection{Adversarial Prototype Fission}

Typical PL frameworks~\cite{yang2018robust, lu2022pmal} with multiple prototypes are usually trained towards a $\max$-based loss function $\mathcal{L}_{max}$ by assigning a sample to its nearest prototype out of multiple~(some include margining closest negative prototype) as follows:
\begin{equation}
\begin{aligned}
    &logit(x) = \max(sim(f(x), p_{(c,1)}),..., sim(f(x), p_{(c,V)})),\\
    &\mathcal{L}_{max}(x) = R_e(\sigma(logit(x)), c) 
\end{aligned}
\end{equation}
where $sim(f(x), p_{(c,i)})$ is the similarity between feature $f(x)$ extracted from sample $x$ and one of $V$ prototypes, $logit(x)$ is the logit of input $x$, $\sigma$ is an activation function and $R_e$ is empirical risk. The similarity term can vary from geometrical similarity, e.g., Mahalanobis similarity or Euclidean similarity, to Triangle similarity, e.g. cosine similarity. In this paper, we use cosine similarity as the basic form.

However, for learn-able prototypes, this $max$ operation usually degrades multiple pre-defined prototype slots to only one effective, by always assigning an incoming sample to the same prototype.

As multiple learnable prototypes are clearly the optimal formation in SSL schemes, we introduce an Adversarial Prototype Fission training strategy to properly train these prototypes while maintaining consistency and diversity. 

\subsubsection{Diversity Modeling:} 
To make sure designed $V$ prototypes are properly mutually dissimilar, a local divergence regularization is included as follows:
\begin{equation}
    \mathcal{L}_{ldiv}(l_{c}) = \sum_{i\neq j} sim(l_{(c,i)},l_{(c,j)})
\end{equation}

To further make sure samples are properly assigned to $V$ per-class prototypes in the training phase, it is mandatory to design a sample branching strategy. Firstly, we compute an Softmax-formed assignment vector $\mathcal{A}(x)$ from each sample $x$ of class $c$ as follows:
\begin{equation}
\begin{aligned}
    &\mathcal{A}(x) = [a(f(x), p_{(c,1)}), ..., a(f(x), p_{(c,V)})],\\
    &a(f(x), p_{(c,i)}) =\frac{\exp{(T\cdot sim(f(x), p_{(c,i)}))}}{\sum_{v}^{V}{\exp{(T\cdot sim(f(x), p_{(c,v)}))}}}
\end{aligned}
\end{equation}
where $T>>1$ is a temperature term. $T$ discriminates the prototype with maximum similarity from other prototypes: $p_{(c,i)}$ with a slightly larger similarity produces an assignment vector with significantly lower entropy. 

Then, for multiple samples $x_i, i \in [1,2, ..., M]$ of class $c$ within a batch, the diversity loss $\mathcal{L}_{div}(x)$ is computed as follows:
\begin{equation}
    \mathcal{L}_{div}(x) = KL([\mathcal{A}(x_1),\mathcal{A}(x_2), ...,\mathcal{A}(x_M)]\|\mathcal{P(\mathcal{A})})
\end{equation}
where $P(\mathcal{A})$ is a prior distribution. The prior distribution is empirically assumed as uniform distribution so that each of $V$ prototypes is activated; as this assumption is weak, the aforementioned temperature $T$ serves as a neutralizer that further alleviates potential negative impact on the original similarity $p_{(c,i)}$.

\subsubsection{Consistency Modeling:} As the diversity modeling encourages mining different sub-categories, the risk of over-focusing on low-level appearances and ignoring shared high-level features remains a challenge. Our consistency-encouraging operation is simple yet effective: in addition to $\max$-based loss function $\mathcal{L}_{max}$, we propose an $average$-based consistency loss function $\mathcal{L}_{cst}$ as follows:
\begin{equation}
\begin{aligned}
    &logit(x,i) = sim(f(x), p_{(c,i)}),\\
    &\mathcal{L}_{cst}(x) = \frac{1}{V}\sum_{i}^{V}(R_e(\sigma(logit(x,i)), c)) 
\end{aligned}
\end{equation}
where $logit(x,i)$ is computed on each prototype $p_{(c,i)}$. This loss function encourages that all prototypes of class $c$ are properly activated by $x$.

\subsection{Overall Loss Function} 

By weighted summing the aforementioned three loss functions, we form the final overall loss function as follows:
\begin{equation}
    \mathcal{L} = (\mathcal{L}_{max} + \lambda_{div} \cdot \mathcal{L}_{div}) + \lambda_{cst} \cdot \mathcal{L}_{cst}
\end{equation}
where $\lambda_{cst}$ and $\lambda_{div}$ are loss weights. It's obvious that $\mathcal{L}$ forms adversarial objectives: $\mathcal{L}_{cst}$ encourages that all of $V$ prototypes are sufficiently activated, while $\mathcal{L}_{div}$ and $\mathcal{L}_{max}$ encourage that one of $V$ prototypes is slightly more activated. Optimizing this $\mathcal{L}$ helps maintain both consistency and diversity. 

Finally, identical loss function $\mathcal{L}$ is used for each supervised sample $x_{sup}$ with a manual label $c$, and each unsupervised sample $x_{unsup}$ with pseudo label $c_{pseudo}$. Pseudo labels are derived from $\sigma(logit(x_{unsup}))$

\section{Experiments}

\subsection{Implementation Details}

The tested methods are all performed on a workstation platform with 128GB RAM and $4\times$ NVIDIA TESLA V100 GPUs with 32GB GPU memory using Ubuntu 18.04 system. The code is implemented with Python 3.7 and PyTorch 1.6.0. 

To verify the effectiveness of our method, we conducted two experiments that combined Prototype Fission with two popular SSL frameworks: ORCA for open-world SSL that formulates novel classes from unlabeled data, and FixMatch for evaluating our contribution to standard SSL. CIFAR-10 and CIFAR-100~\cite{krizhevsky2009learning} are used to evaluate the performance on both a limited number and a larger number of classes.

\subsubsection{Hyper-parameters:} The hyper-parameters across different frameworks are generally aligned with ORCA~\cite{cao22} listed as follows: The backbone feature extractor is adopted as ResNet18~\cite{He2016Deep} pretrained by SimCLR~\cite{chen2020simple}, which is shared across all competing methods. The optimizer is set as SGD with a momentum of 0.9 and the weight decay is set as $5\times 10^{-4}$. The initial learning rate is 0.1 and decays by a factor of 10 at epochs 140 and 180. 

The Prototype Fission-specific hyper-parameters are empirically set as follows: The similarity function $sim(f(x), p_{(c,i)})$ is set as cosine similarity. The number of prototypes $V$ is set as $5$. The weight for diversity loss $\lambda_{div}$ is set as $0.001$ for CIFAR-10 and $0.01$ for CIFAR-100. The weight for consistency loss $\lambda_{cst}$ is set as $0.6$ for CIFAR-10 and $0.3$ for CIFAR-100. The temperature for prototype assigning is set as $T = 10$. 


\subsubsection{Prototype Fission + FixMatch:} We combine Prototype Fission with FixMatch~\cite{sohn2020fixmatch} to validate its effectiveness on typical SSL methods without non-trivial open-set-specific efforts. Specifically, we form FixMatch and PF+FixMatch as follows: 



(1) FixMatch-Sigmoid: This manipulation of FixMatch ditches the Softmax activation function and uses Sigmoid as $\sigma$ instead as a more proper form of FixMatch in open-set settings. Being aware of the existence of OOD samples, we further extend FixMatch to also assign less-confident (controlled by threshold $thr$) pseudo-labeled samples as OOD instead of ignoring them in regular SSLs. For these OOD samples, a binary-cross-entropy-based loss replaces $\mathcal{L}$ as follows:
\begin{equation}
\begin{aligned}
    \mathcal{L}(x_{ood}) = BCE(\sigma(logit(x_{ood})), 0)
\end{aligned}
\end{equation}
In the experimental result part we will show its effectiveness.

(2) PF+FixMatch-Sigmoid: By injecting PF to FixMatch, we firstly replaced its $N$-class fully connected layer with with our $VN$ prototypes. Then we further adapt our cosine similarity $logit(x)$ in $\mathcal{L}_{max}$ and $logit(x,i)$ in $\mathcal{L}_{cst}$ to a tempered Sigmoid activation function as follows:
\begin{equation}
    \begin{aligned}
        &\sigma(logit(x)) = Sigmoid(T \cdot logit(x)-b),\\
        &\sigma(logit(x,i)) = Sigmoid(T \cdot logit(x,i)-b)
    \end{aligned}
\end{equation}
The temperature $T$ is set as $10$ and the offset $b$ is set as $5$ by referring to the Sigmoid curve. This operation empirically facilitates a better convergence compared to the direct usage of cosine similarity. 

The probability threshold $thr$ is set fixed as $0.95$ as in the original FixMatch across four aforementioned variants. In the latter two variants, pseudo-labels with maximum ID probability $\sigma(logit(x))<thr$ are marked as OODs. The batch size is set as 64 for supervised data and 448 for both weak augmentation and strong augmentation unsupervised data. 




\subsubsection{Prototype Fission + ORCA:} We also combine Prototype Fission with ORCA~\cite{cao22} to validate its effectiveness on SSL methods with novel class discovery mechanism. As ORCA forms extra slots for novelties, there is no need for replacing Softmax with Sigmoid. Therefore, we keep full output channels in prototypes as same as ORCA~(10 for CIFAR-10, 100 for CIFAR-100) where former 50\% channels deal with ID samples and the rest slots are used to cluster OOD samples.

The unsupervised loss in ORCA is built upon minimizing cosine similarity between Softmax probabilities of closest pairs in a batch:
\begin{equation}
    \begin{aligned}
        &\mathcal{L}_{p} = \frac{1}{P}\sum_{(i,j)}^{P} -\log<\sigma(logit(x_i)),\sigma(logit(x_j))>
    \end{aligned}
\end{equation}
where $x_i$ and $x_j$ are closest pair (w.r.t. $logit(x_i)$) in the unsupervised data and $P$ is the number of pairs. We adapt our method to this pairwise operation as follows:

(1) The activation function $\sigma$ is extended to a tempered Softmax function:
\begin{equation}
    \begin{aligned}
        &\sigma_{c}(logit(x)) =\frac{\exp{(T\cdot sim(f(x), p_{c}))}}{\sum_{i}^{N}{\exp{(T\cdot sim(f(x), p_{i}))}}}
    \end{aligned}
\end{equation}
where $N$ is the summed number of seen and novel classes. The Softmax function is further poisoned by the ORCA uncertainty $\mu$ in the supervised learning loss.

(2) The unsupervised pairwise loss is further divided into a $max$-based pairwise loss and an $average$-based pairwise loss as follows:
\begin{small}
\begin{equation}
    \begin{aligned}
        &\mathcal{L}_{max}(x) = \frac{1}{P}\sum_{(i,j)}^{P} -\log<\sigma(logit(x_i)),\sigma(logit(x_j))>,\\
        &\mathcal{L}_{cst}(x) = \frac{1}{VP}\sum_{(i,j)}^{P}\sum_{v}^{V} -\log<\sigma(logit(x_i,v)),\sigma(logit(x_j))>
    \end{aligned}
\end{equation}
\end{small}

(3) In addition to ORCA's soft pairwise clustering for pseudo-labeling, hard pseudo-labeling has to be introduced to activate the diversity loss $\mathcal{L}_{div}$. To this end, the pseudo-labeling threshold $thr = 0.95$ is used only for computing $\mathcal{L}_{div}$.
\begin{figure*}[ht]
	\centering
	\centerline{\includegraphics[width=16cm]{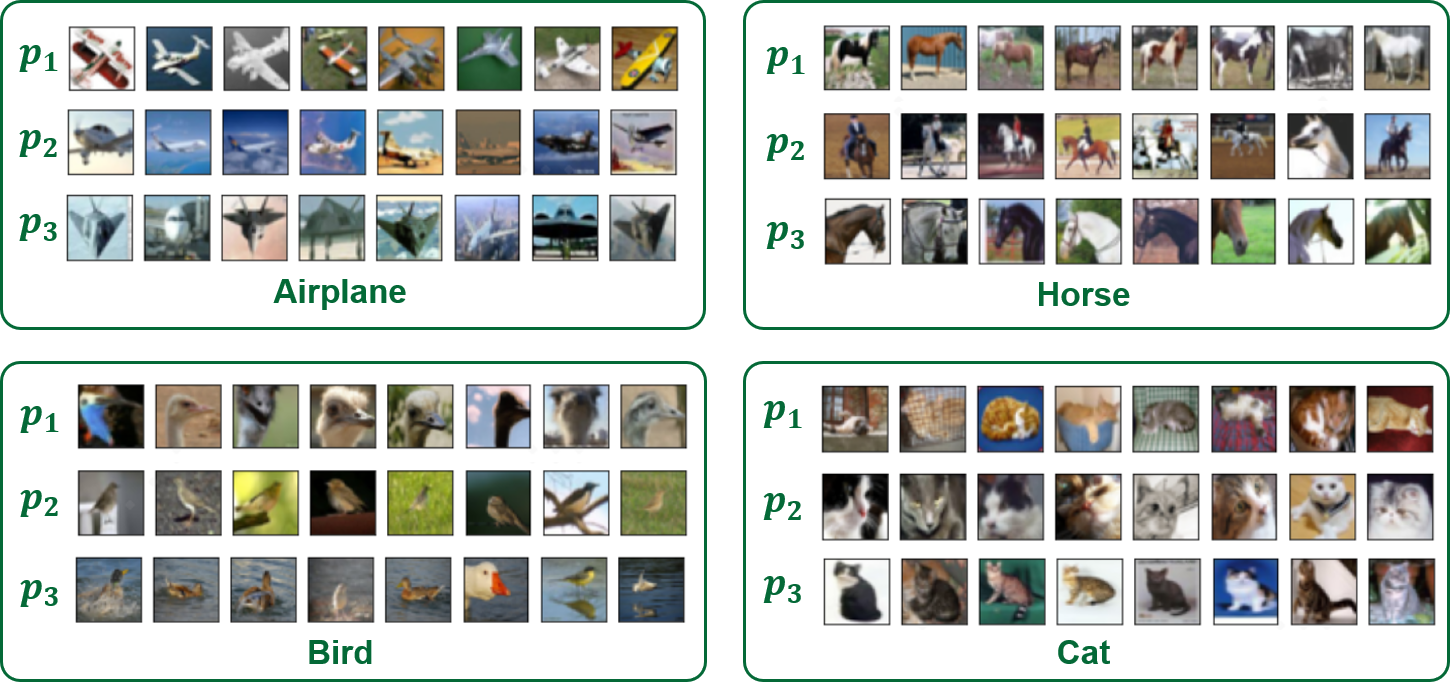}}
	\vspace{-0.0cm}
	\caption{Sub-semantics mined from CIFAR-10 by Prototype Fission. Each prototype $p_{i}$ mined a row of semantically similar CIFAR-10 test images, for example, different prototypes divide the horse class into 'the full-length portrait of a horse', 'human riding a horse' and 'head of a horse'. }
	\label{fig:Visualization}
\end{figure*}



\subsection{Datasets}

We evaluate Prototype Fission on CIFAR-10 and CIFAR-100~\cite{krizhevsky2009learning} datasets. The CIFAR-10 dataset consists of 60,000 32x32 color images in 10 classes, with 6000 images per class. There are 50,000 training images and 10,000 test images. CIFAR-100 extends 10 classes to 100 classes with the same total number of images at 60,000. There are 500 training images and 100 testing images per class. To compare our method to existing works, we evaluated our method on two different ID-OOD splits according to open-world and open-set settings as follows:

\subsubsection{ORCA-like Split:} Firstly, we follow the split scheme of ORCA~\cite{cao22}: The 50,000 training set is first split into two parts: 50\% as seen classes and 50\% as novel classes. Then, 50\% of samples of seen classes are selected as the supervised data, i.e. 2500 samples per seen class for CIFAR-10 and 250 samples per seen class for CIFAR-100. The rest forms the unsupervised data consisting of 12,500 seen class images and 25000 novel class images. The unsupervised data is used for both SSL and testing. This split setting forms 67\% ID-OOD mismatch $num_{unsup\_ood}/ num_{unsup}$ in both the unsupervised data and the test data. The list of seen/novel classes and corresponding image samples exactly followed ORCA practice.


\subsubsection{DS3L-like Split:} To further evaluate the robustness of PF across different ID-OOD mismatch ratios, we follow the split scheme of DS3L~\cite{guo2020safe}. This split also divides the dataset into 50\% of seen classes and 50\% novel classes. The labeled data is set as 400 images per class in CIFAR-10 experiments, and 100 images per class in CIFAR-100. The number of unsupervised images is fixed to 20,000 and the 20,000 images are divided by the distribution mismatch ratio $num_{unsup\_ood}/ num_{unsup}$. In our experiments, the distribution mismatch ratio ranges from 0.1 to 0.7. The test set contains 10,000 images: 1000 per class and all 10 classes included for CIFAR-10, 100 per class, and 100 classes included for CIFAR-100. The list of seen/novel classes and corresponding image samples exactly followed DS3L practice. 

\subsection{Metrics}

\begin{table*}[!htb]
 	\caption{C{\upshape omparing the results on CIFAR-10 and CIFAR-100 using ORCA Metrics. We report the average accuracy over three runs}}
	\centering
	\label{tab:ood}
    \begin{tabular}{lccccccccc}
    \toprule
    & \multicolumn{3}{c}{ \textbf{CIFAR-10}} & \multicolumn{3}{c}{\textbf{CIFAR-100}} & \multicolumn{3}{c}{\textbf{ImageNet-100}} \\
        \textbf{Method}  & \textbf{Seen} & \textbf{Novel} & \textbf{All} & \textbf{Seen} & \textbf{Novel} & \textbf{All} & \textbf{Seen} & \textbf{Novel} & \textbf{All} \\
\midrule
        FixMatch~\cite{sohn2020fixmatch}  &71.5 & 50.4 & 49.5& 39.6 & 23.5 & 20.3 & 65.8 & 36.7 & 34.9 \\ 
        DS3L~\cite{guo2020safe}  & 77.6 & 45.3 & 40.2 & 55.1 & 23.7 & 24 & 71.2 & 32.5 & 30.8 \\ 
        CGDL~\cite{sun2020conditional}  & 72.3 & 44.6  & 39.7 & 49.3 & 22.5 & 23.5 &  67.3 & 33.8 & 31.9 \\ 
        DTC~\cite{han2019learning} & 53.9 & 39.5 & 38.3 & 31.3 & 22.9 & 18.3  & 25.6 & 20.8 & 21.3\\ 
        RankStats~\cite{han2020automatically} & 86.6 & 81.0 & 82.9 & 36.4 & 28.4 & 23.1 & 47.3 & 28.7 & 40.3\\ 
        SimCLR~\cite{chen2020simple}  & 58.3 & 63.4 & 51.7 & 28.6 & 21.1 & 22.3 & 39. & 35.7 & 36.9 \\ 
        ORCA~\cite{cao22}   & 88.2 & 90.4 & 89.7 & 66.9 & 43.0 & 48.1 & 89.1 & 72.1 & 77.8 \\
\midrule
        ORCA~(Our Re-trained)  & 88.1 & 87.7 & 87.8 & 66.5 & 41.0 & 46.6 & 89.4 & 69.1 & 75.5 \\
        PF+ORCA  & \textbf{88.6} & \textbf{88.8} & \textbf{88.7} & 66.0  & \textbf{41.3} &  \textbf{47.9} & \textbf{90.0} & \textbf{74.4} & \textbf{79.6}\\
\bottomrule
    \end{tabular}
\end{table*}

\subsubsection{Accuracy Metrics:} Firstly, The accuracy metrics are reported following ORCA~\cite{cao22} for fair comparison. It is necessary to elaborate on these accuracy terms in ORCA practice:

(1) Seen Acc:  The standard classification accuracy evaluated on the Seen classes subset of the test set. Note that predicting Novel labels (and incorrect Seen labels) on the Seen subset causes accuracy loss yet predicting Seen labels on the Novel subset does not affect this metric.

(2) Novel Acc: The clustering accuracy evaluated on the Novel classes subset of the test set, using the Hungarian algorithm for class matching~\cite{han2019learning}. Note that predicting Seen labels (and incorrect Novel labels) on this subset causes accuracy loss yet predicting Novel labels on the Seen subset does not affect this metric.

(3) All Acc: The clustering accuracy evaluated on images of All classes, using the Hungarian algorithm for class matching~\cite{han2019learning}. This accuracy is computed by averaging over images instead of classes. Note that this All Acc is not a simple combination of Seen Acc and Novel Acc.

\begin{figure*}[ht]
	\centering
	\centerline{\includegraphics[width=16cm]{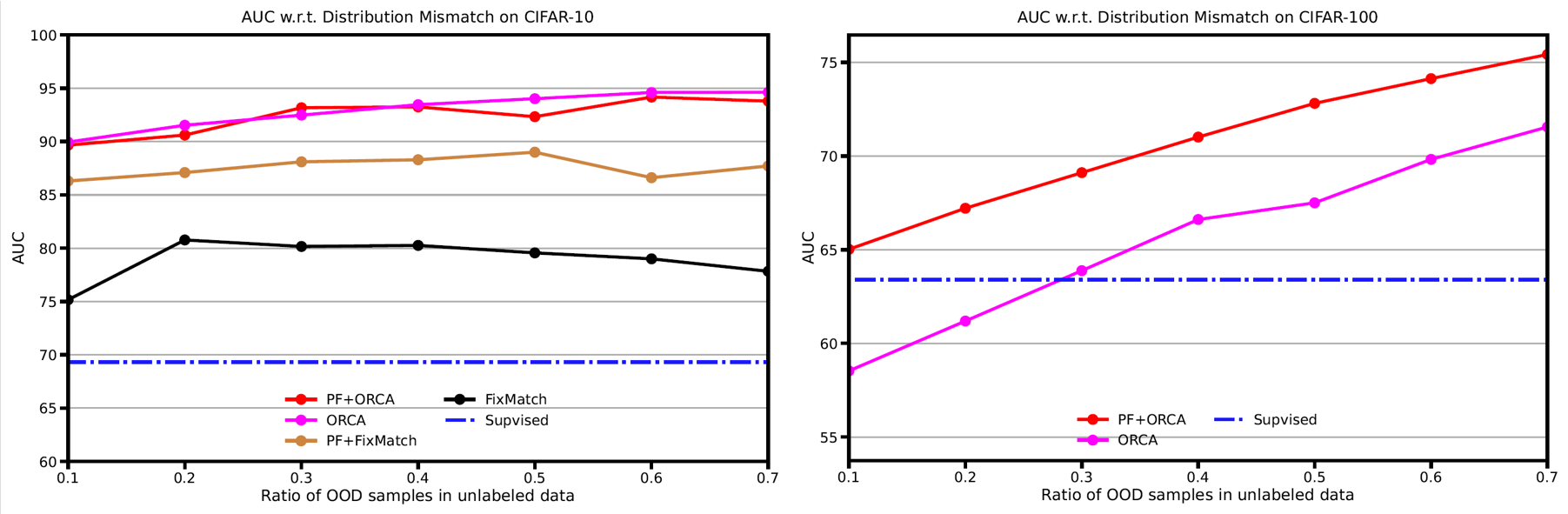}}
	\vspace{-0.0cm}
	\caption{By increasing the ratio of OOD samples in the unsupervised data, Novel-class forming SSL method ORCA enjoys improved ID-OOD discrimination yet regular SSL method FixMatch suffered from degradation despite pseudo-labeling uncertain samples as OOD. By combining our Prototype with both ORCA and FixMatch, a consistent improvement in ID-OOD discrimination is observed from each mismatch ratio.}
	\label{fig:discCurve}
\end{figure*}

\subsubsection{Robustness Metric: } Secondly, following DS3L~\cite{cao22} for a fair comparison, we further adopt the AUC metric to evaluate the confusion between ID and OOD classes. ID-OOD AUC transfers multi-class classification with OOD awareness to a binary classification task by setting $\max_{c}(\sigma_c(logit(x)))$ as ID probability and $1-\max_{c}(\sigma_c(logit(x)))$ as OOD probability. Then, we report the area under curve~(AUC) of this binary task w.r.t. different ID-OOD distribution mismatch ratio in the unsupervised dataset. Similar results were reported as (1-AUC) in DS3L~\cite{guo2020safe}.


\subsection{Results}

\subsubsection{Visualization:} As is observed from Fig. \ref{fig:Visualization}, Prototype Fission divides samples within each class into sub-types according to reasonable rules: sub-types (e.g.,'fighters' out of airplanes), different views (e.g., 'horse head' out of horses, 'birds head' out of the bird and 'cat head' out of cats), different statuses (e.g. 'riding horse' out of horses, 'flying planes' and 'landed planes' out of planes, 'swimming birds' out of birds) and different poses~(e.g., different cat poses). The visualization validated that our Prototype Fission method is functioning as intended. The following experiments further validate that dividing sub-classes is beneficial to both overall accuracy and OOD rejection capability. 

\subsubsection{Classification Accuracy:} We compare our method to ORCA following \textbf{ORCA-like Split} and ORCA metrics to evaluate its accuracy gains as is listed in Tab. \ref{tab:ood}. Note that all third-party methods compared are manipulated by ORCA to be capable of recognizing novel classes so that Hungarian algorithm-based ORCA metrics can be reasonably computed. By injecting Prototype Fission into ORCA, we got $0.92\%$ overall accuracy improvement on CIFAR-10 and $1.33\%$ overall accuracy improvement on CIFAR-100. Note that the original form of FixMatch and PF+FixMatch are not compared to in this experiment due to their incapability of recognizing novel classes. It is also worth noting that we re-trained ORCA using its released code and got slightly inferior results than reported. 

\subsubsection{Robustness to Distribution Mismatch:} We compare our method to existing works following \textbf{DS3L-like Split} to evaluate its robustness to ID-OOD distribution mismatch in the unsupervised data on different ID-OOD mismatch points, as is illustrated in Fig. \ref{fig:discCurve}. In the CIFAR-10 experiment, PF+FixMatch significantly outperformed FixMatch, yet the combination with ORCA showed limited difference; In the CIFAR-100 experiment which is more challenging, PF significantly outperformed ORCA in terms of discriminating OOD samples. 

\begin{table}[!htb]
\begin{small}
	\caption{A{\upshape blation Study}}
	\centering
	\label{tab:ablation}
    \begin{tabular}{lcccccc}
    \toprule
    & \multicolumn{3}{c}{ \textbf{CIFAR-10}} & \multicolumn{3}{c}{\textbf{CIFAR-100}} \\
        \textbf{Parameter} & \textbf{Seen} & \textbf{Novel} & \textbf{All} & \textbf{Seen} & \textbf{Novel} & \textbf{All}   \\
\midrule
    PF+ORCA &  88.60 & 88.81 & 88.74 & 65.99 & 41.34 & 47.92\\
    \midrule
    w/o $p_{(c,i)}$ & 88.06 & 87.71 & 87.82 & 66.45 & 40.98 & 46.59 \\
    w/o $\mathcal{L}_{cst}$ & 88.02 & 77.34 & 76.04 & 70.93 & 35.72 & 40.22 \\
    w/o $\mathcal{L}_{div}$ & 87.80 & 88.51 & 88.27 & 65.53 & 38.81 & 45.87\\ 
\bottomrule
    \end{tabular}
\end{small}
\end{table}

\subsubsection{Ablation Studies: }
In this section, we explore the contribution of each component by removing the component from PF+ORCA. The results are reported in ORCA metrics evaluated on CIFAR-10 and CIFAR-100 as is listed in Tab. \ref{tab:ablation}.

(1) The contribution of multiple prototypes. By removing multiple prototypes $p_{(c,i)}$,  the framework degrades to the original ORCA, and Consistency Modeling and Diversity Modeling are both no longer effective. 

(2) The contribution of Consistency Modeling. By removing the consistency modeling loss $\mathcal{L}_{cst}$, each class is divided into multiple sub-classes with little agreement shared. This causes sub-type learning overly biased to local contents, and therefore ORCA metrics drop by significant margins $\delta All Acc > 5$.

(3) The contribution of Diversity Modeling. By removing the diversity modeling loss $\mathcal{L}_{div}$, multiple prototypes are designed yet not sufficiently utilized. Similar to CPL~\cite{yang2018robust}, in most cases, samples are assigned to the same prototype, and remaining prototypes are bypassed. To some extent, PF degrades to the original ORCA, yet certain gains are still noticed due to the remaining activation of multiple defined prototypes. 


\section{Conclusion}

This work presents Prototype Fission as a novel method for safe Semi-supervised Learning given an open unsupervised dataset and open test dataset. Prototype Fission "closes" ID spaces by dividing class-wise latent spaces into sub-class spaces to make ID spaces more compact, thereby reducing ID-OOD confusion. Prototype Fission can be not only used separately but also combined with existing OOD-aware SSL methods for further performance gain. Extensive experiments showed the effectiveness of our method in open-set and open-world SSL tasks. 

\bibliography{aaai23}

\end{document}